\documentclass[lettersize,journal]{IEEEtran}
\usepackage{amsmath,amsfonts}
\usepackage{algorithmic}
\usepackage{algorithm}
\usepackage{array}
\usepackage[caption=false,font=normalsize,labelfont=sf,textfont=sf]{subfig}
\usepackage{textcomp}
\usepackage{stfloats}
\usepackage{url}
\usepackage{verbatim}
\usepackage{graphicx}
\usepackage{cite}
\usepackage{xfrac}
\usepackage{siunitx}
\usepackage{makecell}
\usepackage{multirow}
\usepackage{soul}
\usepackage{threeparttable} 
\begin{document}

\title{Automotive Object Detection via Learning Sparse Events by Spiking Neurons}

\author{Hu Zhang, Yanchen Li, Luziwei Leng~\IEEEmembership{Member,~IEEE}, Kaiwei Che, Qian Liu, Qinghai Guo, \\
Jianxing Liao, and Ran Cheng~\IEEEmembership{Senior Member,~IEEE}
    \thanks{This work was supported in part by Guangdong Natural Science Funds for Distinguished Young Scholar under Grant 2024B1515020019. (\emph{Hu Zhang and Yanchen Li contributed equally to this work.})  (\emph{Corresponding authors: Luziwei Leng and Ran Cheng.})}
    \thanks{Hu Zhang, Yanchen Li, Kaiwei Che and Ran Cheng are with the Southern University of Science and Technology, Shenzhen 518055, China.
            (e-mail: ranchengcn@gmail.com) 
            }
    \thanks{Luziwei Leng, Qian Liu, Qinghai Guo and Jianxing Liao are with the Advanced Computing and Storage Lab, Huawei Technologies Co., Ltd., Shenzhen 518055, China. (e-mail: lengluziwei@huawei.com)}

}

\maketitle

\begin{abstract}
Event-based sensors, distinguished by their high temporal resolution of \SI{1}{\micro\second} and a dynamic range of \SI{120}{\decibel}, stand out as ideal tools for deployment in fast-paced settings like vehicles and drones. 
Traditional object detection techniques that utilize Artificial Neural Networks (ANNs) face challenges due to the sparse and asynchronous nature of the events these sensors capture. 
In contrast, Spiking Neural Networks (SNNs) offer a promising alternative, providing a temporal representation that is inherently aligned with event-based data. 
This paper explores the unique membrane potential dynamics of SNNs and their ability to modulate sparse events.
We introduce an innovative spike-triggered adaptive threshold mechanism designed for stable training. Building on these insights, we present a specialized spiking feature pyramid network (SpikeFPN) optimized for automotive event-based object detection. 
Comprehensive evaluations demonstrate that SpikeFPN surpasses both traditional SNNs and advanced ANNs enhanced with attention mechanisms. 
Evidently, SpikeFPN achieves a mean Average Precision (mAP) of 0.477 on the {GEN1 Automotive Detection (GAD)} benchmark dataset, marking significant increases over the selected SNN baselines.
Moreover, the efficient design of SpikeFPN ensures robust performance while optimizing computational resources, attributed to its innate sparse computation capabilities.
Source codes are publicly accessible at \url{https://github.com/EMI-Group/spikefpn}.
\end{abstract}


\section{Introduction}
\IEEEPARstart{O}{bject} detection is fundamental in computer vision, with applications ranging from face recognition to vehicle tracking and the identification of smaller objects~\cite{DBLP:journals/ijcv/LiuOWFCLP20, DBLP:journals/pieee/ZouCSGY23}. Traditionally, such tasks rely on frame-based cameras. However, these often produce images that are blurred or poorly exposed, especially in high-speed or challenging lighting conditions, common in scenarios like emergency vehicle detection.
Addressing this gap, the recently advanced Dynamical Vision Sensor (DVS), a.k.a. the event-based sensor~\cite{DBLP:journals/jssc/LichtsteinerPD08, DBLP:journals/jssc/PoschMW11, DBLP:journals/tvlsi/ChenTZC12, DBLP:journals/jssc/Serrano-GotarredonaL13, DBLP:journals/jssc/BrandliBYLD14, DBLP:conf/isscc/SonSKJKSPLPWRLW17}, offers a compelling alternative. 
Unlike traditional sensors, the DVS draws inspiration from retinal functionalities and is adept at registering pixel intensity variations. 
With its remarkable temporal resolution of \SI{1}{\micro\second} and a dynamic range of \SI{120}{\decibel}, it is able to effectively record asynchronous events instigated by shifts in pixel brightness.

Despite the appealing characteristics, the unique nature of DVS also brings challenges. Its event-driven data is sparse and unpredictable, making traditional object detection methods based on Artificial Neural Networks (ANNs) less effective.
While there have been efforts to preprocess such events using various architectures and algorithms~\cite{DBLP:conf/nips/PerotT0MS20, DBLP:journals/tip/LiLZXHT22}, they often come with computational and latency costs.
By contrast, the Spiking Neural Networks (SNNs)~\cite{DBLP:journals/ijns/Ghosh-DastidarA09, roy2019towards} introduce a new paradigm, resembling the brain's functioning by transmitting information through discrete spikes. 
Their inherent efficiency, marked by sparse activations and multiplication-free inferences, makes them ideal for managing the dynamic, sparse data from event-driven sensors.

Recent breakthroughs in direct training methods have enabled more efficient development of deep SNN architectures~\cite{DBLP:journals/neco/ZenkeG18, DBLP:conf/aaai/WuDLZ0S19, DBLP:journals/spm/NeftciMZ19}. These advancements have paved the way to improve SNN performance in image classification tasks~\cite{DBLP:journals/tnn/RathiR23, DBLP:conf/aaai/Zheng00HL21, DBLP:conf/nips/LiGZDHG21, DBLP:conf/iclr/DengLZG22}.
However, when it comes to more intricate vision tasks like automotive object detection~\cite{DBLP:conf/ijcnn/CordoneMT22, 8890199, DBLP:conf/itsc/UlrichBKNFGB22}, SNNs still find themselves trailing behind the established prowess of ANNs. Though significant strides have been made in enhancing ANNs for challenging object detection tasks~\cite{DBLP:journals/tcsv/ChenSPWF22, DBLP:conf/iccv/Brazil019, DBLP:conf/wacv/GongY0PZH21, DBLP:journals/tgrs/LiangGWVKWZ22, DBLP:journals/tip/ChenWPQ21}, directly transposing these refined ANN architectures onto the SNN framework often leads to compromised performance and heightened latency. Such performance deficits are glaring in tasks that necessitate sophisticated neural network adjustments~\cite{DBLP:conf/nips/HagenaarsPC21, DBLP:conf/aaai/KimPNY20, DBLP:conf/ijcnn/CordoneMT22}. Moreover, while current efforts focus on converting ANNs to SNNs~\cite{DBLP:conf/iclr/BuFDDY022, DBLP:conf/icml/LiD0GG21, DBLP:conf/ijcai/DingY0H21, 10.3389/fnins.2017.00682, DBLP:conf/ijcnn/DiehlNB0LP15}, these approaches grapple with inherent challenges. These primarily arise when integrating the dynamic nature of SNNs with the foundational attributes of ANNs, which frequently results in the underutilization of the unique temporal and sparse characteristics inherent to SNNs --- essential for handling event-driven data efficiently. The inherent differences in data representation and processing between the two neural architectures often result in inconsistencies, undermining both the accuracy and computational performance of the neural system.

In response to these challenges, we delve deeper into the temporal dynamics inherent to spiking neurons, aiming to develop an approach characterized by minimal computational requirements and efficient response times, especially crucial for swiftly varying events. Leveraging our findings, we design a tailored {spiking feature pyramid network (SpikeFPN)} for event-based automotive object detection, which harmoniously blends the unique neuronal properties with surrogate gradient training to achieve optimal performance.
Our primary contributions are delineated as follows:
\begin{enumerate}
    \item 
    We explore the inherent temporal dynamics of spiking neurons, particularly focusing on membrane potential dynamics and adaptive membrane thresholds. By retaining the adaptive feedback mechanism and involving the adaptive spiking neuron model in training,
    our comprehensive study not only reveals their intrinsic strengths in managing rapid event modulations but also enhances feature robustness in sparse event data.
    The insights gained ensure stable and efficient training phases, optimizing the network's performance in real-world scenarios.

    \item We design an event-driven SpikeFPN tailored specifically for automotive object detection, leveraging the self-adaptive mechanism inspired by neocortex neuron adaptation.
    This design foundation captures the unique temporal dynamics of spiking neurons, presenting a novel approach to object detection in automotive scenarios. 
    We directly train our SpikeFPN using a surrogate gradient to facilitate model design, with the selected surrogate gradient function optimizing the development process.

    \item 
    Our proposed SpikeFPN demonstrates promising performance, surpassing previous SNN models and advanced ANNs with attention mechanisms in the realms of current event-based automotive object detection.
    It attains a noteworthy mean Average Precision (mAP) of 0.477 on the GEN1 Automotive Detection (GAD) benchmark dataset, marking a substantial advancement by outperforming the selected SNN baselines, exceeding the best baseline by 9.7\%.
    Additionally, our design is a testament to efficiency --- combining a streamlined architecture with impressive accuracy, all while significantly reducing computation costs.
\end{enumerate}

The structure of this paper is outlined as follows: Section~II provides the foundational background relevant to our study; Section~III delves into the intricacies of the proposed SpikeFPN, detailing its architectural components and elaborating on the neural behavior and adaptive strategies employed; Section~IV showcases our experimental approaches, the methodologies adopted, and the corresponding results, affirming the effectiveness of our design; finally, the paper culminates with conclusions drawn in Section~V.

\section{Background}

\subsection{Event-based Object Detection}
Event cameras, possessing outstanding temporal resolution and dynamic range, are highly suited for scenarios demanding rapid object tracking, variable lighting conditions, and minimal latency. 
Nonetheless, the intrinsic dynamism of event camera data conflicts with conventional deep learning paradigms grounded in frame-based methodologies. 
Typically, to counter this incongruence, events are preprocessed to transform into denser representations before network processing. 
Various methodologies, both handcrafted and automated, have been developed for this transformation, encompassing event frames~\cite{DBLP:journals/pami/LagorceOGSB17, DBLP:conf/cvpr/MaquedaL0GS18, DBLP:conf/ebccsp/MoeysCKVDNKD16, DBLP:journals/corr/abs-1809-08625, DBLP:conf/rss/ZhuYCD18, DBLP:conf/cvpr/NguyenDCT19}, time or event-number stacking~\cite{DBLP:conf/cvpr/WangIHY19}, voxel grids~\cite{DBLP:conf/cvpr/ZhuYCD19}, the event queue approach~\cite{DBLP:conf/iccv/TulyakovFKGH19}, LSTM grids~\cite{DBLP:conf/eccv/CanniciCRM20}, and discrete time convolutions~\cite{DBLP:conf/cvpr/ZhangCZCZGL22}. 
For sparse event data, asynchronous convolutions have also been discussed~\cite{DBLP:journals/ral/ScheerlinckBM19, DBLP:conf/eccv/MessikommerGL020}, with the latter notably applied to the {GEN1 Automotive Detection (GAD) dataset acquired with the GEN1 sensor for event-driven automotive object detection tasks}\cite{DBLP:journals/corr/abs-2001-08499}. 
The work by Perot \emph{et al.}\cite{DBLP:conf/nips/PerotT0MS20} emphasized a convolutional LSTM network coupled with a temporal consistency loss, amplifying training outcomes. 
ASTMNet~\cite{DBLP:journals/tip/LiLZXHT22} showcased a progressive adaptive sampling protocol, fusing temporal attention and memory modules, and clocked significant accuracy levels.

\subsection{Deep SNNs for Vision Tasks}
Neuromorphic hardware evolution has increased interest in employing SNNs in challenging vision-based deep learning tasks~\cite{doi:10.1126/science.1254642, DBLP:journals/pieee/FurberGTP14, leng2016spiking, leng2018spiking, 10.3389/fnins.2019.01201, leng2020solving, DBLP:journals/pieee/DaviesWOSGJPR21}.
When combined with event cameras, these neuromorphic ecosystems are expected to offer power efficiency coupled with minimal latency. To this end, previous studies have opened vital avenues of research~\cite{roy2019towards}.

The ANN-to-SNN conversion strategy revolves around approximating real-valued activations with spiking dynamics~\cite{10.3389/fnins.2017.00682, DBLP:journals/corr/abs-1802-02627}. 
This strategy has achieved high accuracy and provides a viable path for training and using SNN models.
Kim \emph{et al.}\cite{DBLP:conf/aaai/KimPNY20} introduced a spiking version of YOLO for object detection and achieved comparable performance on the PASCAL VOC~\cite{DBLP:journals/ijcv/EveringhamGWWZ10} and MS COCO datasets~\cite{DBLP:conf/eccv/LinMBHPRDZ14}, although it required many time steps to converge. Recent endeavors have managed to reduce latency after conversion~\cite{DBLP:conf/iclr/BuFDDY022, DBLP:conf/icml/LiD0GG21}, yet the scalability of these methodologies to more intricate architectures beyond mere image classification is unclear.

Direct training for SNNs, on the other hand, uses surrogate gradient (SG) functions to mimic backpropagation gradients~\cite{DBLP:journals/neco/ZenkeG18, DBLP:conf/nips/ShresthaO18, DBLP:journals/spm/NeftciMZ19}.
Owing to advancements like tailored encoding, SG, and loss function crafting~\cite{DBLP:conf/nips/Zhang020, DBLP:conf/nips/LiGZDHG21, DBLP:conf/iclr/DengLZG22}, directly trained SNNs have been attaining competitive accuracies on strenuous benchmark tasks, demanding minimal simulation steps for convergence.
These milestones have catalyzed their expansion into other event-centric vision tasks, from optical flow assessment~\cite{DBLP:conf/nips/HagenaarsPC21} to video reconstruction~\cite{DBLP:conf/cvpr/0012WCL0022} and object detection~\cite{DBLP:conf/ijcnn/CordoneMT22}.
Recent breakthroughs~\cite{DBLP:conf/nips/CheLZZMCGL22, DBLP:journals/corr/abs-2304-11857, 10535518} underscore that spike-centric differentiable hierarchical searches can significantly improve SNN performance across a spectrum of event-driven vision tasks, like deep stereo and semantic segmentation.

\subsection{Feature Pyramid Networks}
Feature Pyramid Networks (FPNs) are central components of numerous object detection systems, with their capacity to exploit multi-scale feature information enhancing overall system performance~\cite{DBLP:journals/pami/HeZR015, DBLP:conf/cvpr/LinDGHHB17, DBLP:journals/tnn/ZhaoZXW19}. 
The original FPN design was introduced by Lin \emph{et al.}~\cite{DBLP:conf/cvpr/LinDGHHB17}, which built on traditional pyramid methods by efficiently leveraging single-scale image input, forming a foundation for subsequent advancements in the field.

Among these advancements, the FaPN by Huang \emph{et al.}~\cite{DBLP:conf/iccv/HuangLC021} incorporated a feature alignment module, achieving a significant refinement over the conventional FPN design. 
In contrast to the fusion-based approach, the Single Shot Detector (SSD) method, proposed by Liu \emph{et al.}~\cite{DBLP:conf/eccv/LiuAESRFB16}, delegated different stages of feature maps to detect objects of distinct scales. 
Several integrated techniques, such as those found in previous work~\cite{DBLP:journals/pami/RenHG017, DBLP:conf/iccv/HeGDG17, DBLP:conf/cvpr/CaiV18, DBLP:conf/iccv/LinGGHD17, DBLP:journals/corr/abs-1804-02767}, fused features from diverse stages, predominantly using a top-down unidirectional fusion method --- a prevailing FPN fusion mode in modern object detection models.

Liu \emph{et al.}~\cite{DBLP:conf/cvpr/LiuQQSJ18} made strides by introducing the concept of bottom-up secondary bi-directional fusion, shedding light on the potential benefits of bi-directional fusion. 
Further expanding on the capabilities of FPNs, the ASFF~\cite{DBLP:journals/corr/abs-1911-09516} incorporated an attention mechanism, while complex bidirectional fusion techniques~\cite{DBLP:conf/cvpr/GhiasiLL19, DBLP:conf/cvpr/TanPL20} ventured into more intricate fusion dynamics. 
A novel approach was taken as the Recursive-FPN~\cite{DBLP:conf/cvpr/QiaoCY21}, wherein the fused output of the FPN was reintegrated into the backbone, initiating an additional loop and marking new progress. 
Further bridging the gap between conventional neural networks and the emerging field of SNNs, Fu \emph{et al.}~\cite{DBLP:journals/entropy/FuD22} endeavored to combine feature pyramid structures with SNNs.

\section{Method}

This section begins with the representation of event data. Subsequently, it introduces the proposed feature-pyramid-centric spiking neural network supported by a self-adaptive spiking neuron model. 
To clarify the network design, the section describes the fundamental network architecture, highlighting modules such as the primary backbone, and further presents the formulation of the adaptive spiking neuron model intrinsic to the network.
Ultimately, the training strategy for addressing non-derivable neuron models using approximate information is detailed.

\subsection{Event Data Representation}

Traditional cameras capture visual information in the form of images. In contrast, a DVS detects individual pixel changes in their receptive fields, registering changes in luminance and labeling these as ``events''. Typically, an event is recorded in a tuple format \((t,x,y,p)\), where \(t\) represents the timestamp of the event occurrence; \(x\) and \(y\) denote the two-dimensional pixel coordinates where the event is registered; and \(p\) indicates the polarity of the event, reflecting the direction of luminance change.

In order to encode the discrete event data while retaining its temporal nuances, we employ the Stacking Based on Time (SBT) method for event preprocessing as presented by Wang \emph{et al.}~\cite{DBLP:conf/cvpr/WangIHY19}. The SBT approach, recognized for its real-time processing and low computational overhead, has been integrated into the accumulator modules of mainstream event sensors. Beyond mere data compression, the SBT preserves the granular temporal information of the event stream. It amalgamates events into temporally contiguous frames, facilitating the downstream Spiking Neural Network (SNN) to learn from this rich temporal dataset. Within a single stack input into the detection model, the SBT method partitions the event data uniformly based on the time interval $\Delta t$. It then compresses this data to yield a set of frames. Assuming the event data in the stack can be subdivided into $n$ equal time intervals, the value of each pixel in the $i$-th frame is characterized by the aggregated polarity of events as:
\begin{equation}
    P(x,y) = \mathrm{sign}\!\Biggr(\sum_{t\in T} p(x,y,t)\Biggr),
    \label{eq:sbt}
\end{equation}
where \(P\) designates the pixel value at coordinates \((x,y)\); \(t\) represents the timestamp; \(p\) denotes the event's polarity; and {\(T \in\ \bigr[\sfrac{(i-1)\Delta t}{n},\sfrac{i\Delta t}{n}\bigr]\)} specifies the timeframe of events consolidated into a single frame.

While the SBT approach offers substantial utility, alternative event data representation techniques exist, some of which closely resemble our encoding method. Although not integrated into our model, these techniques occasionally offer superior performance in specific contexts. For instance, in scenarios characterized by infrequent events during the consolidation interval, the SBT might yield a frame with notably sparse values. In such cases, the Stacking Based on the number of Events (SBE) method, as introduced by Wang \emph{et al.}~\cite{DBLP:conf/cvpr/WangIHY19}, emerges as a more apt choice. The SBE methodology builds a frame using a predetermined number of events, providing an effective solution to the challenge of sparse event representation. To comprehensively evaluate the influence of various event encoding strategies on object detection tasks, we have compared multiple encoding modalities in the ablation studies detailed in Section~\ref{subsec:abs}.

\subsection{Network Design}
In response to the challenges of event-based object detection, we propose the SpikeFPN --- a spiking neural network built upon the threshold-adaptive spiking unit mechanism with the feature pyramid network architecture as the core principle. The foundational neuron unit, characterized by self-adaptive and spiking attributes, flexibly adapts to intricate scenarios. Coupled with the feature pyramid network module, it provides a robust architecture. A multi-head prediction mechanism layered on top further enhances detection accuracy.
This section discusses the overall network architecture, encapsulating the design intricacies of both the backbone and the feature pyramid module. Subsequently, it illustrates the behavior and propagation mechanisms of the spiking neuron unit incorporated within our approach.

\begin{figure*}[t!]
\centering
\includegraphics[width=1\linewidth]{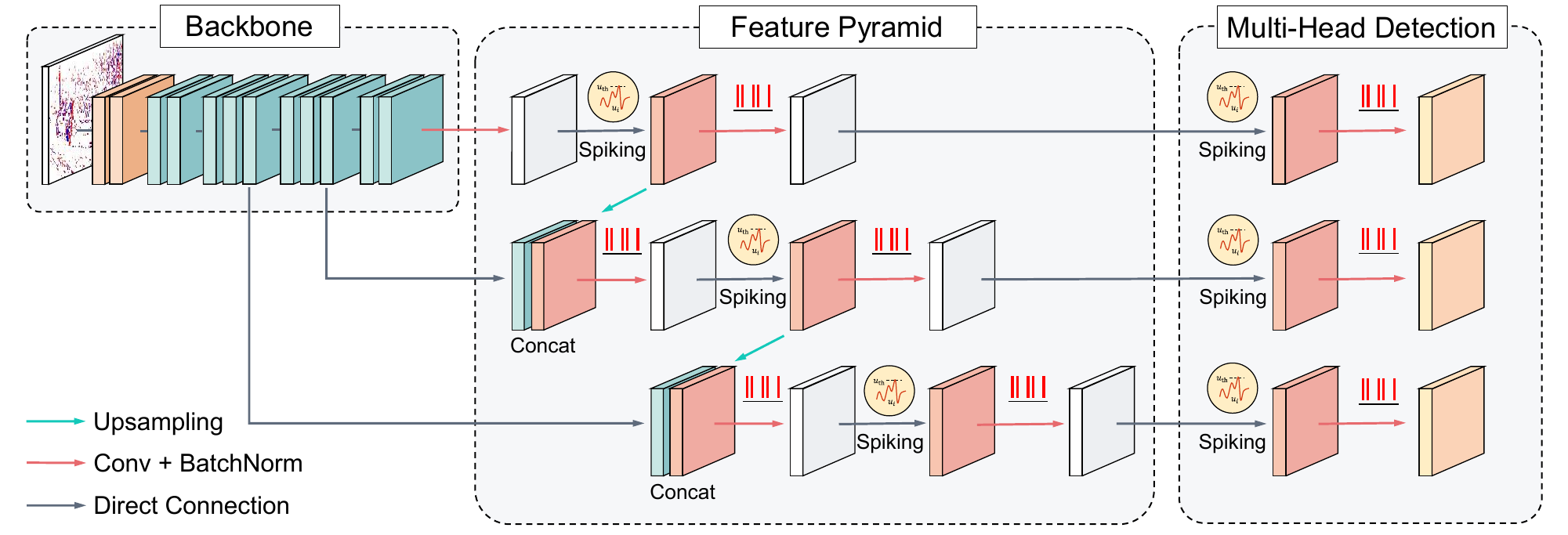}
\caption{Overview of the proposed spiking feature pyramid network. The design encompasses an encoder backbone facilitated by a multi-stage spiking network. Different stages of the backbone contribute to the integrated spiking feature pyramid. Subsequent to this, a multi-head prediction module processes the feature pyramid's output through parallel spiking convolution layers, culminating in the generation of multiple prediction boxes. These are ultimately refined using the non-maximum suppression (NMS) method. Notably, the entire network operates on spike-based computation, allowing the advantage of multiplication-free inference.}
\label{fig:overall}
\end{figure*}

\begin{table}[htbp]
\caption{Detailed architecture settings of the proposed spiking feature pyramid network. The illustration encompasses the encoder backbone, the feature pyramid, and the multi-head prediction module. Here, \(C\) represents the number of classes, while \(K\) signifies the number of anchors.}
\centering
\renewcommand{\arraystretch}{1.2}
    \begin{tabular}{l|c|c}\hline
    \textbf{Module}&\textbf{Layer}&\makecell[c]{\textbf{Output Feature Map}\\\textbf{\(c \times h \times w\)}}\\
    \hline
    \multirow{12}{*}{\makecell[c]{\textbf{Backbone}}}
    &Stem 0 & \(48 \times 128\times 128\) \\
    &Stem 1 & \(96 \times 64\times 64\) \\
    &Cell 0 & \(96 \times 64\times 64\) \\
    &Cell 1 & \(96 \times 64\times 64\) \\
    &Cell 2 & \(192 \times 32\times 32\) \\
    &Cell 3 & \(192 \times 32\times 32\) \\
    &Cell 4 & \(192 \times 32\times 32\) \\
    &Cell 5 & \(384 \times 16\times 16\) \\
    &Cell 6 & \(384 \times 16\times 16\) \\
    &Cell 7 & \(384 \times 16\times 16\) \\
    &Cell 8 & \(768 \times 8\times 8\) \\
    &Cell 9 & \(768 \times 8\times 8\) \\
    \hline
    \multirow{3}{*}{\textbf{Feature Pyramid}}
    &Cell 4 \(\rightarrow\) p1 & \(96\times32\times 32\) \\
    &Cell 7 \(\rightarrow\) p2 & \(192\times16\times 16\) \\
    &Cell 9 \(\rightarrow\) p3 & \(384\times8\times 8\) \\
    \hline
    \multirow{3}{*}{\textbf{Multi-Head Prediction}}
    &p1 \(\rightarrow\) d1 & \(K\times(C+5)\times32\times 32\) \\
    &p2 \(\rightarrow\) d2 & \(K\times(C+5)\times16\times 16\) \\
    &p3 \(\rightarrow\) d3 & \(K\times(C+5)\times8\times 8\) \\
    \hline
    \end{tabular}
\label{tab:SpikeFPN}
\end{table}

\subsubsection{Network Architecture}
The network architecture adopts a feature pyramid module which is based on a downsampling scheme, employing foundational cell-based units to construct the encoding backbone. 
Within such a design, the primary focus shares similarities with the thoughts of YOLOv3~\cite{DBLP:journals/corr/abs-1804-02767}, which draws on the pyramid feature map handling process, with small-sized feature maps being employed to detect large-sized objects while large-sized feature maps detect small-sized objects.
Spiking feature maps, derived from the concluding three stages of the backbone, combine to create the spiking feature pyramid --- a focal element of the primary network design as depicted in Fig.~\ref{fig:overall}.
This component plays a pivotal role in feature extraction. A comprehensive explanation of the backbone's output feature map, in conjunction with the entire network, can be found in Table~\ref{tab:SpikeFPN}.
The first two spiking stem layers of the architecture facilitate double downsampling of the feature map. This is succeeded by 10 spiking cells, which together forge a four-stage downsampling hierarchy. 
Outputs from the terminal cells across stages 2, 3, and 4 are fed into the feature pyramid module. 
The full architecture achieves a maximum downsampling ratio of 32.
In the following, we further explain the design elements of the backbone, complemented by the subsequent spiking feature pyramid network.

\begin{figure*}[htbp]
\centering
\includegraphics[width=1\linewidth]{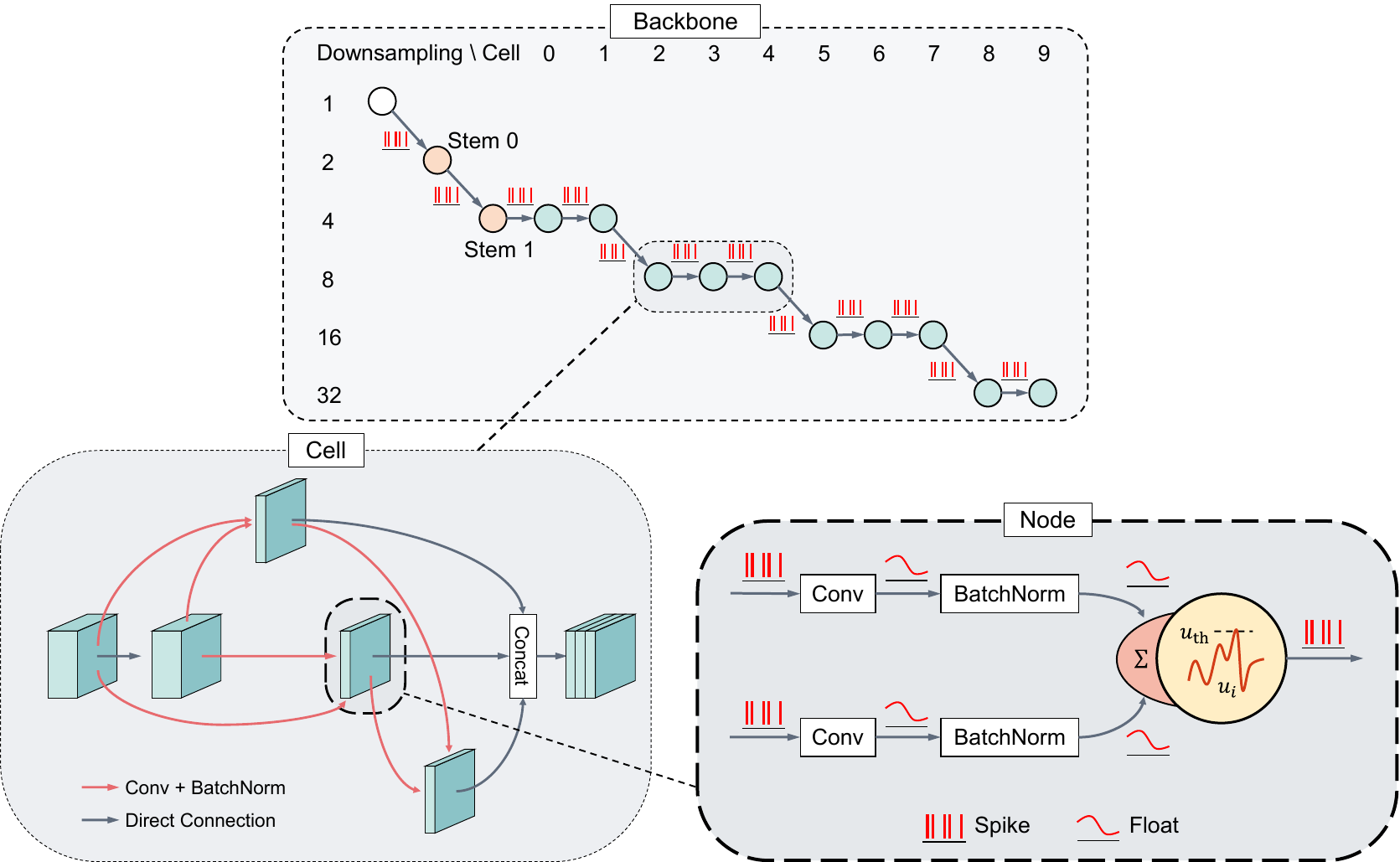}
\caption{Overview of the proposed encoder backbone. The structure adopts a multi-stage downsampling scheme, incorporating multiple cells interlinked via directed acyclic graphs. The vertical annotations on the left delineate the downsampling scale associated with each cell structure, while the numbers atop the left side correspond to the cell's subscript, aligning with the detailed architecture showcased in Table~\ref{tab:SpikeFPN}. Two spiking convolution layers, termed as stem layers, facilitate initial channel variations. The unique cell connection topology recurs across different layers. Operations within each cell, executed at the node level, accumulate on the membrane potential tier, culminating in the form of spikes.}
\label{fig:backbone}
\end{figure*}

As illustrated in Fig.~\ref{fig:backbone}, the encoder backbone adopts a multi-stage downsampling strategy. This architecture consists of two initial spiking stem layers, followed by a sequence of ten spiking cells.
Through the downsampling process, the spatial resolution of the feature map is reduced by half, while the channel size is doubled. 
Each spiking stem layer comprises a convolution layer, a batch normalization layer (which can be integrated into convolution weights during test inference, as suggested by~\cite{DBLP:conf/icml/IoffeS15}), and a spiking activation function. 
These layers are specially designed to propel initial feature extraction and channel modulation.
The local design of the architecture is based on the inspiration of SpikeDHS~\cite{DBLP:conf/nips/CheLZZMCGL22}. We adopt tailored cells and nodes to compose this architecture.
In SpikeFPN, a cell is defined as a repeating unit with an internally directed acyclic graph structure consisting of nodes, each of which receives the results of the previous two cells (if any) as input; a node is a pre-defined block of spiking neural network which is the basic component of a cell, the specific architecture of the node and cell is represented as in the underside of Fig.~\ref{fig:backbone}.
Each spiking cell is composed of three spiking nodes, with this arrangement being repeated across multiple layers.
The initial pair of nodes acquire inputs from the preceding two cells, while the third node receives input from its immediate two predecessors.
The feature maps across all nodes maintain a consistent size. Their respective outputs are converged, forming the collective output of the cell.
Within each node, operations stemming from various levels aggregate at the membrane potential level before the spiking activation generation.
The interconnections within the spiking cell are heuristically determined to harmonize inference speed with accuracy.

After processing through the backbone module, fundamental features are discerned. 
However, subsequent operations are also necessary to achieve the anticipated results. The lower spatial resolution features from the pyramid are upsampled via the nearest interpolation method, which doubles both width and height dimensions.
These are subsequently concatenated with feature maps (of congruent spatial dimensions) generated from the backbone.
Within the overall network architecture, the feature pyramid module enhances the features, synergizing the backbone components, thereby promoting good adaptability for intricate tasks at hand. 
These enhanced features are processed by a \(1 \times 1\) convolution layer, followed by a batch normalization layer and a spike activation function, constituting the succeeding module of the feature pyramid. 
The spiking feature pyramid then interfaces with a multi-head prediction module through convolution layers, batch normalization layers, and spike activation functions. 
The resultant features are processed by a \(1 \times 1\) convolution, yielding floating-point outcomes with the shape of \(K \times (C+5) \times h \times w \).
Here, \(h\) and \(w\)  represent the height and width of the corresponding feature maps, respectively; \(K\), the number of anchors, is set to \(3\); \(C\) denotes the number of classes; and the additional 5 accounts for four bounding box coordinate predictions and one confidence prediction.
The method of non-maximum suppression (NMS)~\cite{DBLP:conf/cvpr/RedmonF17,DBLP:conf/cvpr/GirshickDDM14} is employed to finalize the bounding boxes.

\subsubsection{Neural Behavior and Adaptation}
In the realm of SNNs, the choice of the basic neuron unit emerges as a pivotal aspect of network design. 
In event-based real-time scenarios, adaptability to the conveyed information is a core concern. 
As such, we adopt an adaptively tunable spiking neuron model. The iterative Leaky Integrate-and-Fire (LIF) neuron~\cite{DBLP:conf/aaai/WuDLZ0S19} is adopted as the fundamental unit, which is then augmented with a firing threshold-adaptive mechanism, enhancing its representational capabilities for event data, as discussed below.

Leveraging its inherent temporal dynamics and sparse activation, the LIF neuron model's propagation pattern can be articulated as:
\begin{subequations}
\begin{align}
    u_i^{(t,n)} &= \tau\cdot u_i^{(t-1,n)}\cdot\bigr(1-y_i^{(t-1,n)}\bigr) + I_i^{(t,n)}, \\[1.5ex]
    I_i^{(t,n)} &= \sum_jw_{ji}\cdot y_j^{(t, n-1)}, 
\end{align}
\label{eq1:LIF}
\end{subequations}
with the spike generation process formulated with:
\begin{equation}
    y_i^{(t,n)} = H\bigr(u_i^{(t,n)}-u_\text{th}\bigr) =
    \Biggr\{
    \begin{aligned}
    1,~&\text{if}\ u_i^{(t,n)} \geq u_\text{th}, \\[1.5ex]
    0,~&\text{otherwise.}
    \end{aligned}
\label{eq2:spike}
\end{equation}
where the \(u_i^{(t,n)}\) represents the membrane potential of the neuron \(i\) of the \(n\)-th layer at time \(t\), \(\tau\) is a constant which represents the membrane time attenuation factor consequently configures the membrane leakage of the neuron, \(I_i^{(t,n)}\) is the influx currency, which is essentially the weighted sum of the spiking activation \(y_j^{(t, n-1)}\) generated from the connected neurons from the previous layer \(n-1\). The spiking activation \(y_i^{(t,n)}\) is defined by a Heaviside function \(H(\cdot)\) which is calculated to be \(1\) only when \(u_i^{(t,n)}\) reaches a membrane threshold \(u_\text{th}\), otherwise remains \(0\).

Drawing inspiration from the adaptive behaviors of neurons in the neocortex~\cite{gerstner2014neuronal, gouwens2018systematic}, our design incorporates the self-adaptive mechanism into the LIF spiking neuron. 
This self-adaptive mechanism has demonstrated adaptability across various experimental setups and applications. 
Liu \emph{et al.}\cite{DBLP:journals/tamd/LiuLWCWS23} and Chen \emph{et al.}\cite{DBLP:journals/ijon/ChenMFX22} have enhanced spiking neurons with adaptive properties, verifying the effectiveness of these adaptations from both biological and application perspectives. 
Furthermore, Yang \emph{et al.}\cite{10323199}, Chen \emph{et al.}\cite{10229179}, and Wei \emph{et al.}\cite{DBLP:conf/iccv/WeiZQBZC23} developed dynamic mechanisms for adjusting the membrane threshold over time.
Meanwhile, Sun \emph{et al.}\cite{DBLP:journals/tamd/SunCYCXYG24} explored making the membrane threshold a trainable parameter, synergistically training it with synaptic weights to enhance model performance.

Our enhancement module incorporates an adaptive threshold mechanism previously utilized in recurrent SNNs for temporal sequence learning~\cite{DBLP:conf/nips/BellecSSL018, bellec2020solution}, enabling longer memory retention. 
Unlike previous models, our approach maintains a mechanism where the membrane threshold is influenced by outputs while simplifying the overall model. 
This provides specific feedback properties that differ from those where the threshold changes only over time. 
To maximize this module's effectiveness, we made its hyper-parameters trainable, allowing it to actively participate in the training and modulation processes. We strategically apply this module at the first layer of SpikeFPN to balance flexibility and stability.
Specifically, the adaptive threshold mechanism's propagation pattern can be described as:
\begin{subequations}\label{eq2}
\begin{align}
    y^{(t)} &= H\bigr(u^{(t)}-A_{\text{th}}^{(t)}\bigr), \\[1.5ex]
    A_{\text{th}}^{(t)} &= u_\text{th}+\beta\cdot a^{(t)}, \\[1.5ex]
    a^{(t)} &= \tau_{a}\cdot a^{(t-1)}+y^{(t-1)},
\end{align}
\end{subequations}
where \(H(\cdot)\) is the Heaviside step function. \(A_{\text{th}}^{(t)}\) signifies the modifiable threshold at time \(t\). Meanwhile, \(a^{(t)}\) represents the evolving threshold increment, which is modulated by the neuron's spiking history. 
Here, \(\beta\) is a scaling coefficient, and \(\tau_{a}\) is a trainable time constant for \(a\). 

Crucially, the adjustable threshold \(A_{\text{th}}\) dynamically influences the spiking rate. When faced with intensive input, this threshold rises, thereby inhibiting the neuron's firing propensity. 
On the other hand, sparse input leads to a decrease in this threshold, making the neuron more prone to firing. 
By examining the limits of this adaptive mechanism under various input scenarios, we can discern its bounds. 
Under conditions where the neuron remains consistently inactive, as \(t\) approaches infinity, \(a^{(t)}\) converges to 0. This implies that the lowest bound for \(A_{\text{th}}^{(t)}\) is the base threshold \(u_\text{th}\). However, in a contrasting scenario where the neuron is incessantly active (with initial conditions \(y^{(0)} = 1\) and \(a^{(0)} = 0\)), the evolution of \(a^{(t)}\) can be portrayed as:
\begin{equation}
a^{(t)} = \sum_{i=1}^{t} \tau_{a}^{(i-1)},
\label{eq:eqac}
\end{equation}
and its upper limit, as \(t\) grows indefinitely, is determined to be \(\sfrac{1}{(1-\tau_{a})}\). Hence, the adaptive threshold \(A_{\text{th}}\) spans from \(u_\text{th}\) to \(u_\text{th}+\sfrac{\beta}{(1-\tau_{a})}\). 
This event-driven adaptivity augments the network's resilience and stability, especially in the face of swift alterations in event density. 

\subsection{Network Training}
To capitalize on the temporal dynamics of spiking neurons, the initial obstacle is the intricacy of training SNNs. This challenge stems from the discontinuous nature of spike generation, such that the absence of a gradient renders backpropagation unfeasible.

Early strides in the application of deep SNNs to machine learning tasks were made by transferring parameters from ANN-trained models to corresponding SNNs~\cite{DBLP:conf/ijcnn/DiehlNB0LP15,10.3389/fnins.2017.00682}. 
However, this method is confined to feed-forward architectures devoid of recurrency, making it suitable only for tasks with limited temporal dependencies. 
To overcome this limitation, we embrace the surrogate gradient-based direct training method. 
This approach retains the discontinuous activation mechanism for forward propagation while replacing the derivative of the spike-generating function (see Eq.~\ref{eq2:spike}) with an approximation. 
This substitution enables backpropagation to proceed effectively.
During this training phase, temporal neural dynamics, as detailed in Eq.~\eqref{eq1:LIF}, are meticulously modeled, and the resulting errors are backpropagated through time.
Consequently, surrogate gradient approximations not only facilitate the training of spiking neurons while considering their temporal intricacies but are also inherently apt for tasks with pronounced temporal dependencies, like the continuous object detection challenge we address. 
As a stand-in for the gradient of the spiking function, we utilize the \(\mathrm{Dspike}\) surrogate function~\cite{DBLP:conf/nips/LiGZDHG21}, given by:
\begin{subequations}\label{eq3:Dspike}
\begin{align}
\mathrm{Dspike}(u) &= a\cdot\tanh\bigr(b\cdot(u-c)\bigr) + d,&0 \leq u \leq 1, \\[1.5ex]
\tanh(x) &= \frac{e^x - e^{-x}}{e^x + e^{-x}},
\end{align}
\end{subequations}
where \(u\) denotes the membrane potential. The parameters \((a, b, c, d)\) modify the intrinsic hyperbolic tangent function, \(\tanh\), ensuring its output remains bounded within \([0, 1]\), while varying in shape.
The parameter \(b\), termed the temperature factor, regulates the smoothness of the function. 
In our chosen variant of the \(\mathrm{Dspike}\) function, adjustments ensure that \(\mathrm{Dspike}(0) = 0\) and \(\mathrm{Dspike}(1) = 1\). 
By selecting different hyper-parameters, the \(\mathrm{Dspike}\) function can represent a broader range of surrogate gradients. As it encompasses a family of more stable spike functions, the estimation of gradients remains within a desirable range. 
During model design, only minimal adjustments to the hyper-parameters are required to achieve favorable outcomes. 
This flexibility provides valuable insights for adapting the model to various scenarios. 
With a wide array of surrogate functions at our disposal, we can effectively cater to diverse training needs.

\section{Experiments}

This section delves into a comprehensive examination of our proposed SpikeFPN, underscoring its advantages through rigorous experimentation. 
Initially, we detail the training regimen of SpikeFPN on the GEN1 Automotive Detection (GAD) dataset~\cite{DBLP:journals/corr/abs-2001-08499}, encompassing facets such as data preparation, hyper-parameter optimization, and comparative performance analysis. 
Subsequently, validation experiments of SpikeFPN on the N-CARS dataset~\cite{DBLP:conf/cvpr/SironiBBLB18} are provided to demonstrate the generalizability of the model on different automotive tasks.
Afterwards, ablation studies elucidate the design decisions underpinning our model --- highlighting the nuances of event data preprocessing and architectural considerations.
Finally, capitalizing on the inherent sparse propagation properties of SNNs, we present a comparative analysis of SpikeFPN's computational efficiency and energy consumption.

\subsection{Experiment on GAD}
\label{subsec:impl_details}

A comprehensive procedure encompassing data preparation, training, and testing has been executed on the GAD dataset, yielding noteworthy results. This section delineates the entire experimental procedure, beginning with the details of the input dataset and its encoding settings. This is followed by a discussion on the requisite hyper-parameters and performance evaluation metrics used during the training and testing phases, culminating in an evaluation of the outcomes.

\subsubsection{Input Data Representation}
The GAD dataset, obtained using a Prophesee GEN1 ATIS sensor, is an expansive event-based automotive object detection collection. It consists of over 39 hours of automotive recordings (with a resolution of 304\(\times\)240). Alongside approximately 255,000 manually labeled bounding boxes at \SI{1}{\hertz}, \SI{2}{\hertz}, and \SI{4}{\hertz} denoting cars and pedestrians, the data is divided into training, validation, and test sets. The recordings are segmented into 60-second intervals, with each set containing 1460, 429, and 470 videos, respectively. During this experiment's training and testing phases, the data partitioning in GAD was meticulously followed to ensure accurate representation of results.

In real-world settings, the event data from the sensor is sequential and varies in length. For a more efficient implementation, we processed the variable-length event data through fixed-length segmentation. After encoding the event data, \(S \times C\) frames preceding the label were created to be fed into the network for each sample in the following training. Here, \(S\) represents the number of stacks from the SBT framing process, and \(C\) is the number of frames in each stack. This frame-compression approach corresponds to what is presented in Eq.~\eqref{eq:sbt}. With \(\Delta t = \SI{60}{\milli\second}\), \(n = C = 3\), and \(S = 3\), the segmented event data is transformed into tensor-form data frames, preserving the time information between frames. The minimum temporal resolution here is \(T = \sfrac{\Delta t}{n} = \SI{20}{\milli\second}\).

\subsubsection{Hyper-Parameters Setting}
For the training phase, we employed the AdamW~\cite{DBLP:conf/iclr/LoshchilovH19} optimizer with an initial learning rate of \(0.001\) and a weight decay of \(0.0005\). Pre-training was deemed unnecessary; hence, all models underwent direct training for 30 epochs with a batch size of 32. A warm-up policy was applied using the first epoch's learning rate, which initially starts at a nominal value and then gradually ascends to the prescribed learning rate. The LIF and ALIF neurons have a membrane time constant \(\tau\) of 0.2 and a membrane threshold \(u_\text{th}\) of 0.3. The temperature \(b\) of the \(\mathrm{Dspike}\) function is set at 3. For the ALIF neuron, \(\beta\) is 0.07, while \(\tau_a\) is a learnable parameter constrained to [0.2, 0.4], initialized at 0.3. To assess the network's training error and gauge the model's performance, we opted for the mAP, setting the prediction score to 0.3 as the accuracy metric. For a more comprehensive evaluation, we employed two mAP-based metrics: mAP\(_{50}\) and mAP\(_{50:95}\). The former, mAP\(_{50}\), represents the widely-accepted mean Average Precision with an overlap threshold of 0.5~\cite{DBLP:conf/eccv/LinMBHPRDZ14, DBLP:journals/ijcv/EveringhamGWWZ10}. In contrast, mAP\(_{50:95}\) serves as an alternative metric, indicating the mean Average Precision across 10 IoU ranges ([.50:.05:.95]).

\begin{table*}[t]
\caption{{Comparison of the two mean
Average Precision metrics and the network parameter quantities for different types of networks on the GEN1 Automotive Detection dataset.}}
\label{tab:gad-compare}
\centering
\setlength{\tabcolsep}{17pt}
\renewcommand{\arraystretch}{1.2}
\begin{threeparttable}
\begin{tabular}{l|c c c c c}
\hline
\textbf{Model} & \textbf{Network Type} & \textbf{Params} (M) & \textbf{Time Steps} & \textbf{mAP\(_{50}\)} & \textbf{mAP\(_{50:95}\)} \\
\hline
SparseConv~\cite{DBLP:conf/eccv/MessikommerGL020}& ANN& 133& -& 0.149& - \\
Events-RetinaNet~\cite{DBLP:conf/nips/PerotT0MS20}& ANN& 33& -& 0.340& -\\
E2Vid-RetinaNet~\cite{DBLP:conf/nips/PerotT0MS20}& ANN& 44& -& 0.270& - \\
RED~\cite{DBLP:conf/nips/PerotT0MS20}& ANN& 24& -& 0.400 & - \\
Gray-RetinaNet~\cite{DBLP:conf/nips/PerotT0MS20}& ANN& 33& -& 0.440 & - \\
ASTMNet~\cite{DBLP:journals/tip/LiLZXHT22}& ANN& -& -& 0.467& -\\
\hline
VGG-11 + SDD~\cite{DBLP:conf/ijcnn/CordoneMT22}& SNN& 13& 5& 0.37\tnote{*} & 0.174 \\
MobileNet-64 + SSD~\cite{DBLP:conf/ijcnn/CordoneMT22}& SNN& 24& 5& 0.35\tnote{*} & 0.147 \\
DenseNet121-24 + SSD~\cite{DBLP:conf/ijcnn/CordoneMT22}& SNN& 8& 5& 0.38\tnote{*} & 0.189\\
SpikeFPN (ours) & SNN & 22 & 3 & \textbf{0.477}& \textbf{0.223} \\
\hline
\end{tabular}
\begin{tablenotes}
  \footnotesize
  \item[*] Provided by the authors.
\end{tablenotes}
\end{threeparttable}
\end{table*}

\begin{figure}[ht]
\centering
\includegraphics[width=1\linewidth]{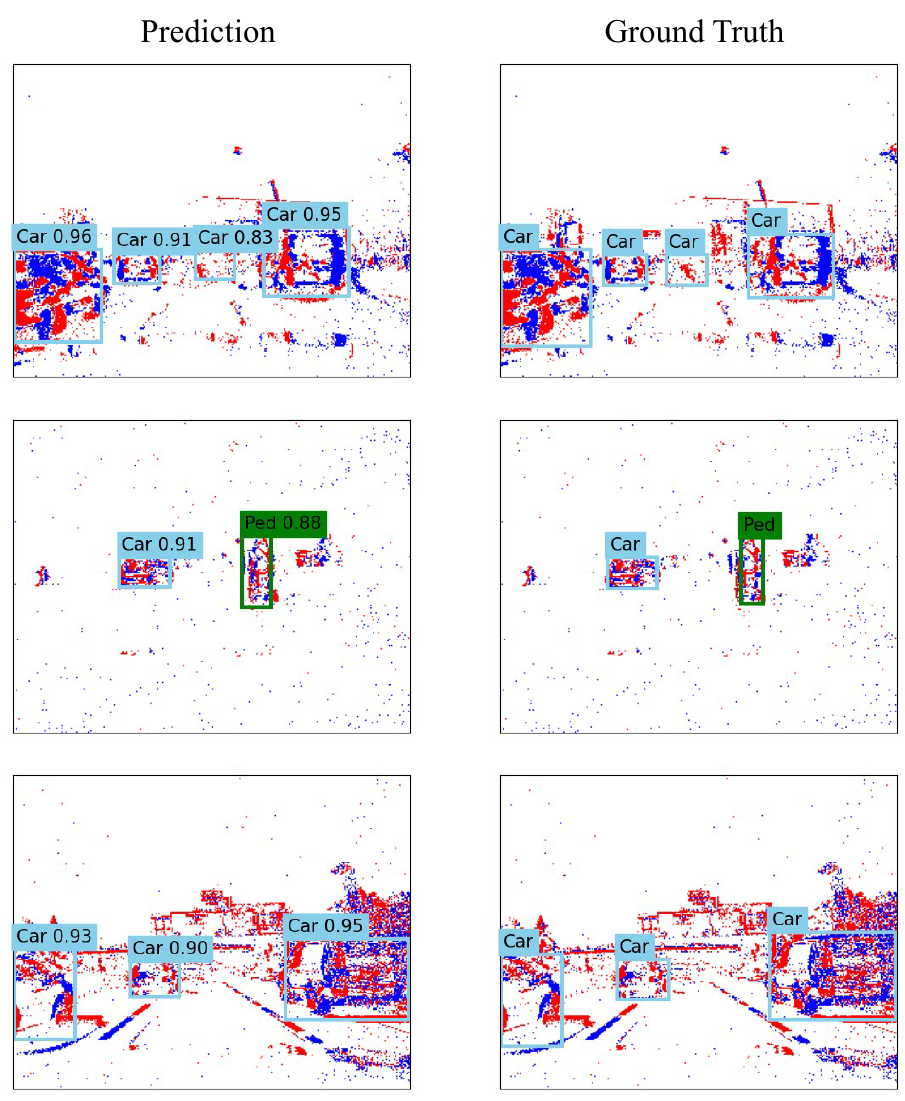}
\caption{{The prediction results of SpikeFPN (left column) and the corresponding ground truth (right column). The ``Ped" in each cell stands for the category of pedestrian.}}
\label{fig:gad-result}
\end{figure}

\subsubsection{Results}
Throughout the training phase, the final stack output from the model serves as the prediction result. This approach ensures the preservation and accumulation of temporal information, crucial for the model's learning process and subsequent calculation of evaluation metrics. For clarity, we juxtaposed our methods against various SNN and ANN models tested on the GAD dataset. These comparative results are tabulated in Table~\ref{tab:gad-compare}, with values for other models sourced from existing literature.

\begin{figure}[b]
\centering
\includegraphics[width=1\linewidth]{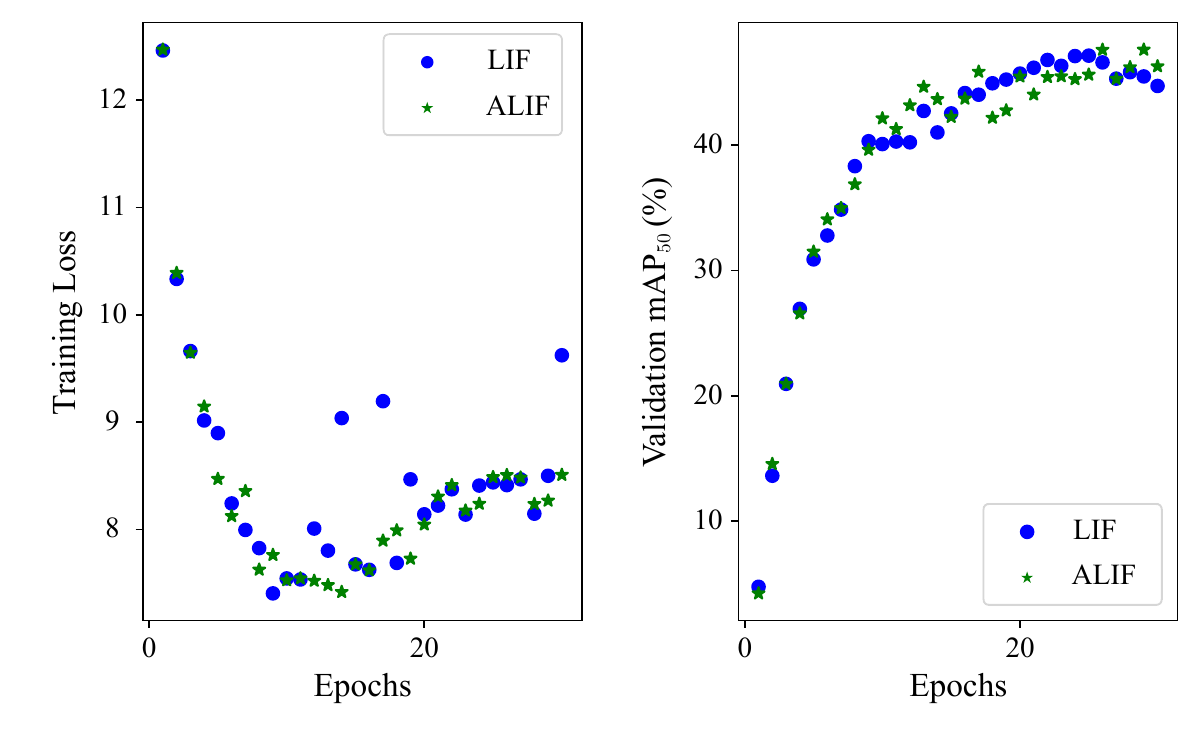}
\caption{Comparisons of training losses and validation mAPs of SNNs using LIF and ALIF neurons for the first layer.}\label{fig:LIF_ALIF}
\end{figure}

From an SNN perspective, our proposed SpikeFPN markedly outperforms three preceding models discussed in~\cite{DBLP:conf/ijcnn/CordoneMT22}. It exhibits an improvement ranging from 9.7\% to 12.7\% for mAP\(_{50}\) and between 3.4\% to 7.6\% for mAP\(_{50:95}\). Remarkably, this is achieved with fewer requisite steps to convergence. An examination of the training loss between SNNs, especially when comparing LIF and ALIF neurons in the first layer (depicted in Fig.~\ref{fig:LIF_ALIF}), reveals that the ALIF neuron fosters a more stable training trajectory than its LIF counterpart. Moreover, our SpikeFPN's streamlined architecture modestly outpaces the ASTMNet~\cite{DBLP:journals/tip/LiLZXHT22}, a state-of-the-art ANN incorporating attention mechanisms and bespoke adaptive sampling event preprocessing schemes.

To assess our model's real-time capabilities, we continuously input the entire test split into the network. This process mirrors the length of steps, generating sequential labels. The average inference velocity for a single stack stands at 54 frames per second when operating on a single GPU. Such outcomes underscore both the efficacy of the SpikeFPN and the broader promise of SNNs in handling dynamic events coupled with sparse propagation. For further illustration, a selection of our model's prediction results can be perused in Fig.~\ref{fig:gad-result}.

\begin{figure*}[t!]
\centering
\includegraphics[width=\linewidth]{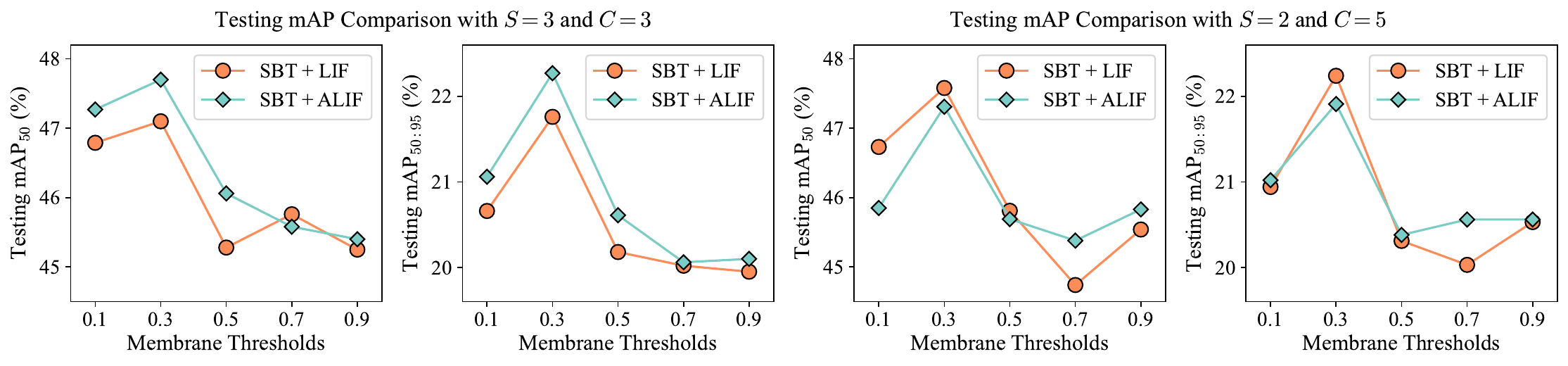}
\caption{The performance comparison of the ``SBT + LIF" versus ``SBT + ALIF" design solutions on the GEN1 Automotive Detection testing dataset with different membrane thresholds.}
\label{fig:map-on-thresholds}
\end{figure*}

\subsection{Experiment on N-CARS}
In this subsection, to verify the generalizability of the model proposed in our work, the N-CARS dataset~\cite{DBLP:conf/cvpr/SironiBBLB18} is adopted to serve as an additional validation experiment for SpikeFPN.
The dataset information and the experimental setup are first presented, followed by the experimental results and their corresponding analysis.

The N-CARS dataset is proposed based on an event-based vehicle recognition task, which falls into a type of binary classification.
In an automotive task scenario, vehicle recognition is one of the major tasks, verifying the performance of SpikeFPN on the N-CARS dataset helps to determine the real-world performance of the model, as well as its generalizability.

\begin{table}[htb]
\caption{Comparison of SpikeFPN accuracy on N-CARS test dataset versus existing baselines. The ``AUC" stands for ``Area Under the Curve".}
\label{tab:ncars}
\centering
\renewcommand{\arraystretch}{1.2}
\setlength{\tabcolsep}{17pt}
\begin{threeparttable}
\begin{tabular}{l|c|c}
\hline
\textbf{Model} & \textbf{Accuracy} & \textbf{AUC Score} \\
\hline
H-First~\cite{DBLP:journals/pami/OrchardMEPTB15} & 0.561\tnote{*} & 0.408\tnote{*} \\
HOTS~\cite{DBLP:journals/pami/LagorceOGSB17} & 0.624\tnote{*} & 0.568\tnote{*} \\
Gabor-SNN~\cite{DBLP:conf/cvpr/SironiBBLB18} & 0.789\tnote{*} & 0.735\tnote{*} \\
HATS~\cite{DBLP:conf/cvpr/SironiBBLB18} & \textbf{0.902}\tnote{*} & \textbf{0.945}\tnote{*} \\
\hline
SpikeFPN (ours) & 0.8694 & 0.8689 \\
\hline
\end{tabular}
\begin{tablenotes}
  \footnotesize
  \item[*] Provided by~\cite{DBLP:conf/cvpr/SironiBBLB18}.
\end{tablenotes}
\end{threeparttable}
\end{table}

In terms of dataset processing, 20\% of the training dataset is divided for validation, the same as the GAD dataset, the SBT method is adopted as preprocessing, whereas the minimum temporal resolution \(T=\sfrac{\Delta t}{n}\) is set to \SI{10}{\milli\second}. The number of frames within each stack (\(C\)) is set to 1 with varied numbers of stacks \(S \in [1, 10]\) depending on the data as the N-CARS max time \SI{100}{\milli\second} for each piece of data. The learning rate is set to 0.0001 to ensure the stability of the training process.

During the training process, the best performing outcome on the validation set is selected as the final weights yielding the testing accuracy and Area Under the Curve (AUC) score as shown in Table~\ref{tab:ncars}. As can be seen from the comparison in the table, the SpikeFPN model proposed in this work, although not specifically designed for classification tasks, still outperforms some of the baselines in the N-CARS dataset and is capable of obtaining comparable performance.

\subsection{Ablation Study}

In this subsection, we conduct ablation experiments from two aspects: the event coding paired with a basic neuron computation scheme, and the structure of the network itself. Through these, we aim to highlight the enhancements the proposed SpikeFPN brings to the table.

\subsubsection{Event Input Configuration}
Our goal here is to gauge the influence of varying preprocessing methods and input configurations on model performance. To this end, we juxtapose results obtained using SBE and SBT, both with LIF neurons. For each method, two input configurations, which are defined by different numbers of stacks (\(S\)) and frames within each stack (\(C\)), are employed: \(S=3\), \(C=3\) and \(S=2\), \(C=5\). With the SBT method, events within \SI{20}{\milli\second} are housed in each frame, while the SBE method's frame encapsulates 5000 events. Additionally, we introduce the Adaptive-LIF (ALIF) neuron to the initial layer in the SBT process, facilitating a comparison of neuron types across various input configurations. The findings are tabulated in Table~\ref{tab:different_method}. 

\begin{table}[htb]
\caption{Results with different input configurations. \(S\) and \(C\)  represent the number of stacks and the number of frames in each stack respectively.}
\label{tab:different_method}
\centering
\setlength{\tabcolsep}{12pt}
\renewcommand{\arraystretch}{1.2}
\begin{tabular}{l|c|c|c|c}
\hline
\textbf{Method} & \(S\) & \(C\) & \textbf{mAP\(_{50}\)} & \textbf{mAP\(_{50:95}\)} \\
\hline
SBE + LIF& 2& 5& 0.4356 &0.1970 \\
SBE + LIF& 3& 3& 0.4324 &0.1944 \\
\hline
SBT + LIF& 2& 5& 0.4758& 0.2224 \\
SBT + LIF& 3& 3& 0.4710& 0.2176 \\
\hline
SBT + ALIF& 2& 5& 0.4731& 0.2191 \\
SBT + ALIF& 3& 3& \textbf{0.4770}& \textbf{0.2227} \\
\hline
\end{tabular}
\end{table}

As seen from the table~\ref{tab:different_method}, it is difficult to guarantee that ``SBT + ALIF" is always the optimal due to the relatively complex correlation between the hyper-parameters, (e.g., at \(S=2,C=5\)), however, in our proposed design, this method is demonstrated experimentally to be the better solution in most cases.
Further experimental analyses are provided as shown in Fig.~\ref{fig:map-on-thresholds}. It can be seen that when selecting \(S=3,C=3\), both metrics mAP\(_{50}\) and mAP\(_{50:95}\) are better with ``SBT + ALIF" than ``SBT + LIF" in most cases;
when selecting \(S=2,C=5\), while ``SBT + ALIF" performs slightly worse in the mAP\(_{50}\) metric, it also shows better performance on mAP\(_{50:95}\) in the majority of situations.

\begin{table}[htb]
\caption{Results with different hyperparameter settings of the adaptive threshold mechanism.}
\label{tab:hyper}
\centering
\setlength{\tabcolsep}{8pt}
\renewcommand{\arraystretch}{1.2}
\begin{tabular}{l|c|c|c|c}
\hline
\textbf{State of \(\tau_a\)} & \textbf{Initial \(\tau_a\)} & \(\beta\) & \textbf{mAP\(_{50}\)} & \textbf{mAP\(_{50:95}\)} \\
\hline
non-trainable & 0.3 & 0.07 & 0.470 & 0.219 \\
trainable & 0.3 & 0.09 & 0.471 & 0.220 \\
trainable & 0.3 & 0.05 & 0.473 & 0.217 \\
trainable & 0.3 & 0.07 & \textbf{0.477} & \textbf{0.223} \\
\hline
\end{tabular}
\end{table}

Preliminary observations indicate that the LIF neuron, when coupled with the SBE process, does not fare as well in terms of accuracy compared to its performance with the SBT process. Interestingly, the configurations \(S=3,C=3\) underperform relative to \(S=2,C=5\) for both SBE and SBT. However, when SBT is paired with the ALIF neuron, the performance is optimized at \(S=3,C=3\). In essence, the SBT process outperforms SBE in our model setup, and the ALIF neuron, when presented with a larger stack number, better modulates its threshold in the SBT process.

\begin{figure}[b]
\centering
\includegraphics[width=0.90\linewidth]{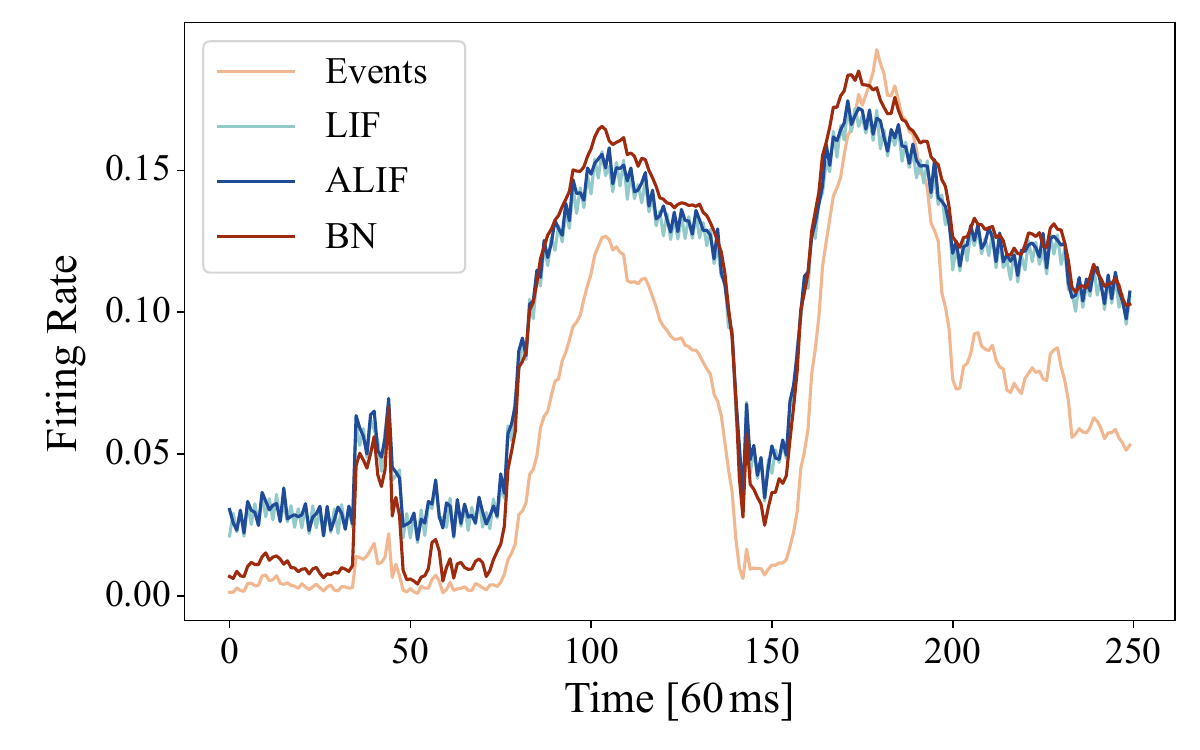}
\caption{Illustration of a sample sequence depicting the neuron firing rates alongside the first layer's activation for LIF \(\tau =0.2\), ALIF \(\tau=0.2\), and binary neurons (denoted as ``BN'' in the legend). Each point on the ``Events'' curve represents the average event density of an SBT stack over a duration of \SI{60}{\milli\second}. The layer activation is derived by averaging the resulting spiking feature map.}
\label{fig:alif_rate}    
\end{figure}

\subsubsection{ALIF Performance}
We further delve into the sensitivity of the adaptive threshold in ALIF concerning hyper-parameters and the merits of training \(\tau_a\). Our findings, summarized in Table~\ref{tab:hyper}, unequivocally show the robustness of performance across a spectrum of \(\beta\) values. Additionally, networks furnished with a trainable \(\tau_a\) consistently outshine those with fixed counterparts.

According to Eq.~\eqref{eq1:LIF}, the membrane time constant, \(\tau\), dictates the proportion of the prior time step's membrane potential to retain. This attenuated accumulation addresses the sparse propagation issue. Fig.~\ref{fig:alif_rate} depicts a sequence of input event density compared to the first layer activation with different \(\tau\) values. Layers with LIF neurons function akin to low-pass filters, especially when event density experiences sharp variations. In contrast, layers with binary neurons (where \(\tau = 0\)) tend to mirror their input closely, displaying synchronous changes. An exploration into the effect of the membrane time constant was undertaken, testing \(\tau\) values from 0 to 1 at 0.1 intervals. With a set random seed for LIF neurons, mAP\(_{50}\) values at \(\tau\) of 0, 0.1, 0.2, and 0.3 were 0.4252, 0.4455, 0.4710, and 0.4696, respectively. This emphasizes the membrane potential constant's impact on overall accuracy. Retaining prior time step data, attenuated during temporal accumulation, proves beneficial. To corroborate the findings, additional experiments with three random seeds were executed for varying \(\tau\) values. The resultant mAP\(_{50}\) values were averaged and presented in Fig.~\ref{fig:alif_scatter}, underscoring the advantages of self-adaptive thresholds during event encoding.

\begin{figure*}[t!]
\centering
\includegraphics[width=0.95\linewidth]{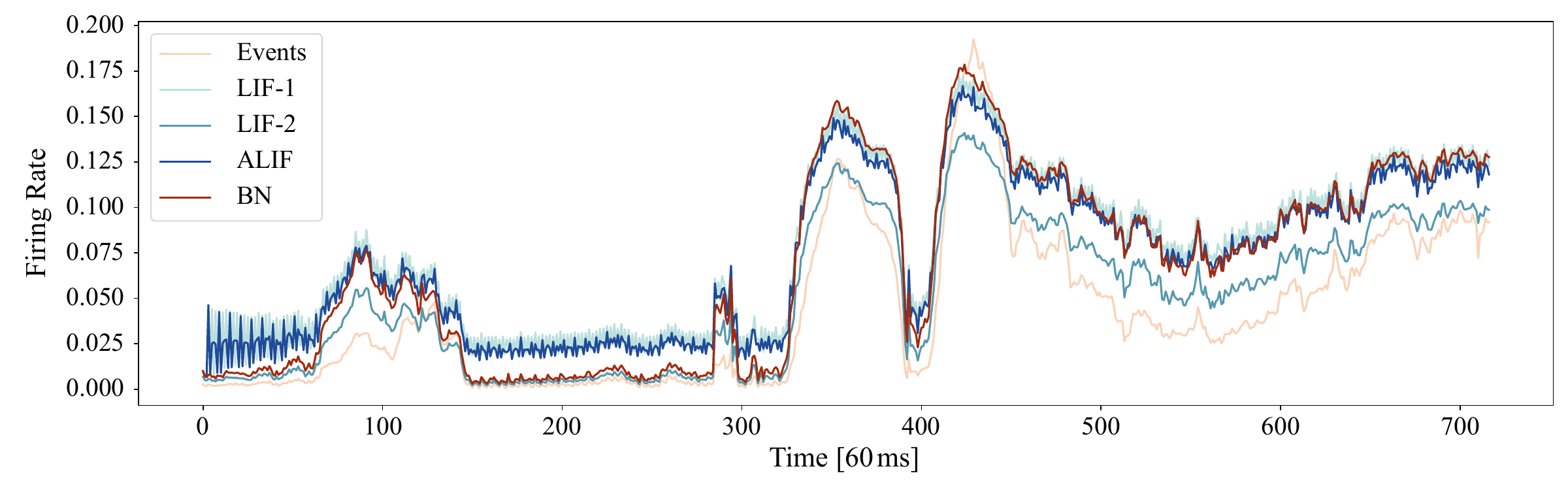}
\caption{Comparison of event density and layer firing rate across LIF, ALIF, and binary neurons (denoted as ``BN'' in the legend). LIF-1 and LIF-2 denote LIF neurons with threshold values set to 0.3 and 0.4, respectively. Each data point on the ``Events'' curve represents the mean event density of an SBT stack spanning \SI{60}{\milli\second}. Layer activation is determined by averaging the spiking feature map.}
\label{fig:sparsity_compare}
\end{figure*}

\begin{figure}[b]
\centering
\includegraphics[width=0.85\linewidth]{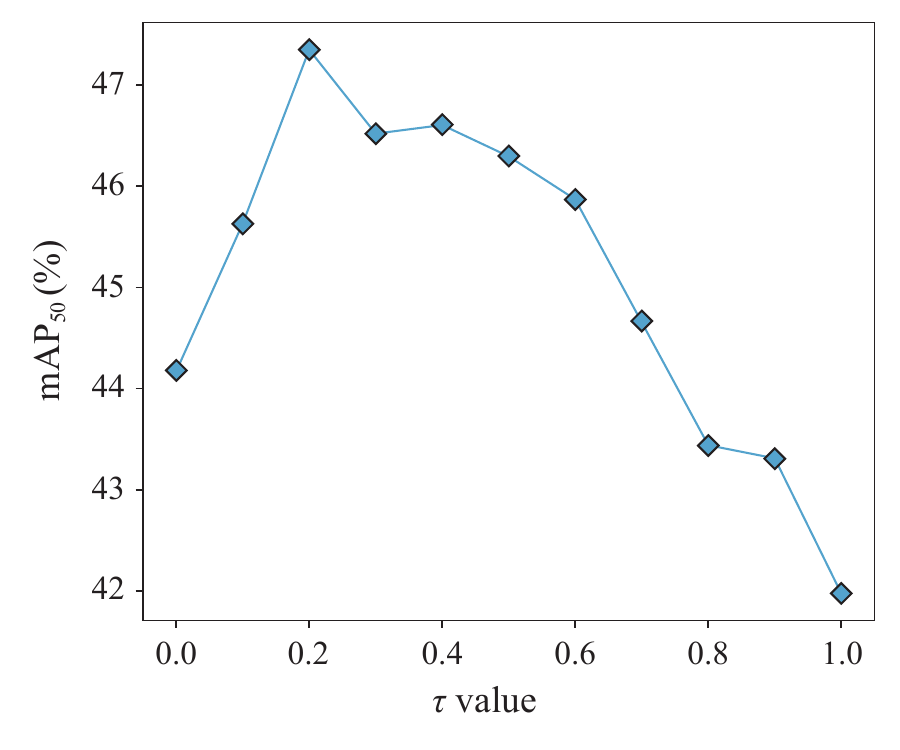}
\caption{A graph detailing the mean average precision for LIF neurons at varying \(\tau\) values. The results, averaged across three distinct experiments with varied random seeds, emphasize the influence of the membrane time constant on performance.}
\label{fig:alif_scatter}
\end{figure}

\begin{figure}[ht]
\centering
\includegraphics[width=0.8\linewidth]{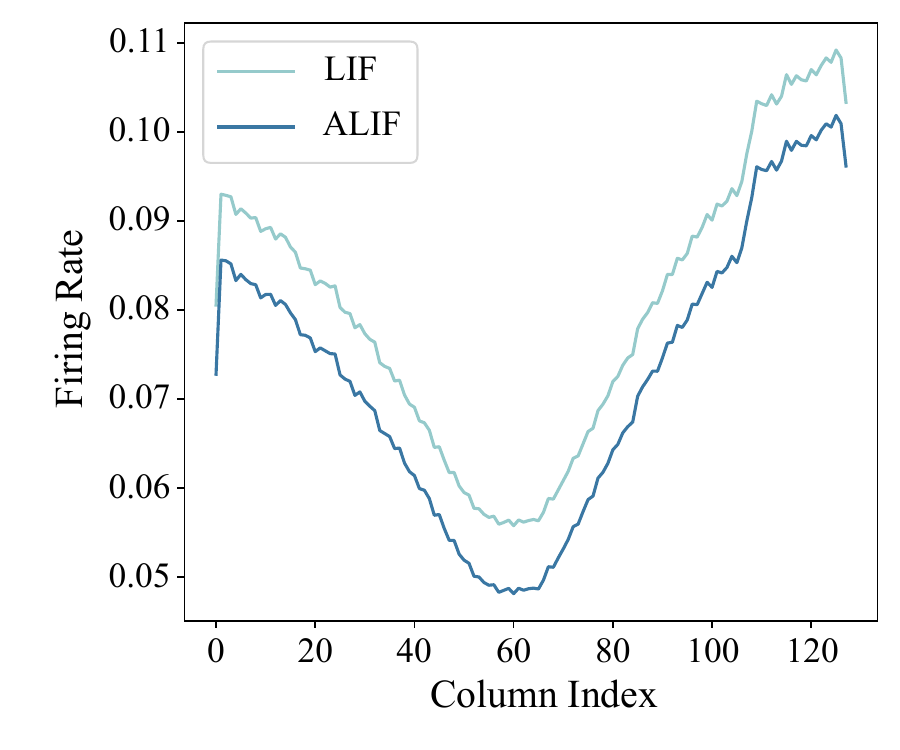}
\caption{Averaged column firing rate of the first layer on the GAD test set for LIF and ALIF neurons. Data was pooled from four experiments, each initiated with a different random seed. The mean mAP\(_{50}\) values for LIF and ALIF neurons stand at \(0.4662\pm0.0029\) and \(0.4738\pm0.0034\), respectively. The V-shaped pattern reflects fewer events from distant objects, particularly at the valley.}
\label{fig:lif_spike}
\end{figure}

When using different random seeds, the network with the first layer utilizing ALIF neurons consistently surpassed networks with LIF neurons at 0.3 and 0.4 thresholds, as detailed in Section~\ref{subsec:impl_details}. This accentuates the adaptive threshold's capability in managing dynamic events. In the context of firing rate, the ALIF neuron leads to decreased layer firing rate, as seen in Fig.~\ref{fig:lif_spike}, emphasizing the merits of self-adjusting thresholds in event encoding.

\begin{table}[b]
\caption{Results on the {GAD} dataset when applying different neuron models on the first and all layers of the network.}
\label{tab:act_func_compare}
\centering
\setlength{\tabcolsep}{12.5pt}
\renewcommand{\arraystretch}{1.2}
\begin{tabular}{l|c|c|c}
\hline 
\textbf{Neuron} & \textbf{Layer}& \textbf{Threshold} & \textbf{mAP\(_{50}\)} \\ 
\hline
LIF & All/First & 0.3 & 0.470 \(\pm\) 0.001 \\
LIF & First & 0.4 & 0.470 \\
\hline
ALIF & All & 0.3 & 0.400 \\
ALIF & First & 0.3 & \textbf{0.476 \(\pm\) 0.002}  \\
\hline
\end{tabular}
\end{table}

In pursuit of optimal neuron placement for the adjustable threshold, we compared the performance of models employing LIF and adaptive threshold LIF (ALIF) neurons. Experiments were conducted using ALIF neurons across the entire network and exclusively on the first layer, with subsequent layers using LIF neurons. During these trials, we treated $\tau_a$ as a trainable variable, consistent across layers. Outcomes are summarized in Table~\ref{tab:act_func_compare}. However, integrating ALIF neurons throughout the network led to subpar outcomes. Over-reliance on adjustable thresholds might infuse excessive flexibility, thereby hampering training precision. For clarity, Fig.~\ref{fig:sparsity_compare} presents an input event stream sequence by event density, juxtaposed with the first layer's activation using distinct neuron models. The LIF layer with \(u_\text{th} = 0.4\) exhibits the least activation, whereas the binary neuron layer with \(u_\text{th} = 0.3\) shows significant fluctuations. In comparison, both LIF and ALIF layers provide a moderating effect over input variations. The ALIF layer's activation resides between the LIF layers' threshold bounds, attributed to the adaptive threshold properties, emphasizing ALIF's potential in decreasing layer firing rate and enhancing accuracy.

\begin{table}[ht]
\caption{Detailed architecture of the Manually constructed spike-based ResNet, including the encoder backbone of ResNet-18, the feature pyramid and the multi-head prediction module. \(C\) denotes the number of classes and \(K\) denotes the number of anchors.}
\centering
\renewcommand{\arraystretch}{1.2}
    \begin{tabular}{l|c|c}\hline
    \textbf{Module}&\textbf{Layer}&\makecell[c]{\textbf{Output Feature Map}\\\textbf{\(c \times h \times w\)}}\\
    \hline
    \multirow{10}{*}{\makecell[c]{\textbf{Backbone}}}
    &Stem 0 & \(80\times128\times128\) \\
    &Stem 1 & \(80\times64\times64\) \\
    &Layer 1-1 & \(80\times64\times64\) \\
    &Layer 1-2 & \(80\times64\times64\) \\
    &Layer 2-1 & \(160\times32\times32\) \\
    &Layer 2-2 & \(160\times32\times32\) \\
    &Layer 3-1 & \(320\times16\times16\) \\
    &Layer 3-2 & \(320\times16\times16\) \\
    &Layer 4-1 & \(640\times8\times8\) \\
    &Layer 4-2 & \(640\times8\times8\) \\
    \hline
    \multirow{3}{*}{\textbf{Feature Pyramid}}
    &Layer 2-2 \(\rightarrow\) p1 & \(80\times32\times32\) \\
    &Layer 3-2 \(\rightarrow\) p2 & \(160\times16\times16\) \\
    &Layer 4-2 \(\rightarrow\) p3 & \(320\times8\times8\) \\
    \hline
    \multirow{3}{*}{\textbf{Multi-Head Prediction}}
    &p1 \(\rightarrow\) d1 & \(K\times(C+5)\times32\times 32\) \\
    &p2 \(\rightarrow\) d2 & \(K\times(C+5)\times16\times 16\) \\
    &p3 \(\rightarrow\) d3 & \(K\times(C+5)\times8\times 8\) \\
    \hline
    \end{tabular}
\label{tab:ResNet_SNN}
\end{table}

\subsubsection{Spiking ResNet}
\label{subsec:abs}
In this ablation study segment, we designed a spike-based ResNet-18 encoder backbone. Distinctive features include variable initial channel sizes, incorporation of the feature pyramid, a multi-head prediction module, and substitution of all activation functions with the Heaviside function \(H(\cdot)\), referenced in Eq.~\eqref{eq2:spike}. This network, adhering to a 4-stage downsampling blueprint, adopts its residual block from~\cite{DBLP:conf/nips/FangYCHMT21, DBLP:conf/iclr/DengLZG22}. Outputs from the last blocks of stages 2, 3, and 4 are relayed to the feature pyramid. The network's maximum downsampling ratio stands at 32. The architecture specifics, when initiated with 80 channels, are tabulated in Table~\ref{tab:ResNet_SNN}. Assessing our encoder's design efficacy, we paralleled it with this variant, given its analogous downsampling approach. Different channel sizes were evaluated to represent diverse model capacities, with findings consolidated in Table~\ref{tab:backbone_compare}. Remarkably, when juxtaposed under similar model parameters, ResNet (initialized with 80 channels) marginally underperforms against SpikeFPN. A plausible explanation could be SpikeFPN's extensive intra and inter-cell connections, fostering gradient and information flow throughout the SNN training.

\begin{table}[ht]
\caption{{Results on GAD dataset with different backbone architectures. The acronym ``ICS" stands for the initial channel size.}}
\label{tab:backbone_compare}
\centering
\renewcommand{\arraystretch}{1.2}
\begin{tabular}{l|c|c|c|c}
\hline 
\textbf{Architecture} & \textbf{ICS} & \textbf{Model Size} (M) & \textbf{mAP\(_{50}\)} & \textbf{mAP\(_{50:95}\)} \\
\hline
ResNet18 &64 &13.34 &0.4163 &0.1799 \\
ResNet18 &80 &20.82 &0.4208 &0.1896 \\
ResNet18 &96 &29.97 &0.4429 &0.1992 \\
\hline
SpikeFPN (ours) & 48 & 21.63 &\textbf{0.4770} &\textbf{0.2227} \\
\hline
\end{tabular}
\end{table}

\subsection{Network Firing Rate and Computation Cost}
SNNs inherently harness the potential of spike-based sparse computations and eschew multiplicative inferences, ensuring a pronounced computational edge over their ANN counterparts, which are predicated on dense matrix multiplication. To elucidate this, we compared SpikeFPN's computational efficiency against existing ANN and SNN benchmarks.

\begin{figure}[b]
\centering
\includegraphics[width=0.95\linewidth]{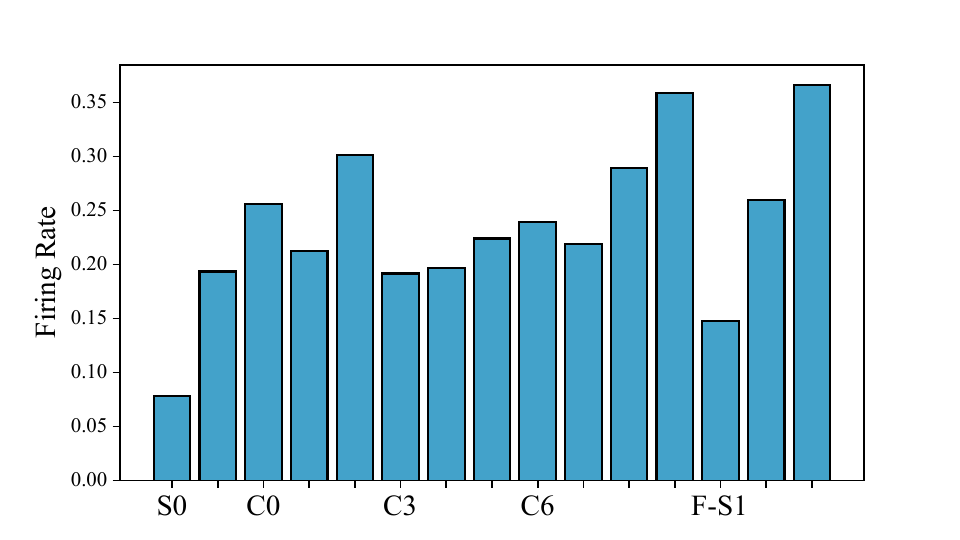}
\caption{The layer firing rates of SpikeFPN. The network exhibits an overall sparse activity with varying degrees across layers, with the lowest firing rate at the first layer due to highly sparse input event streams.}
\label{fig:fig5}
\end{figure}

As illustrated in Fig.~\ref{fig:fig5}, SpikeFPN is characterized by varied layer firing rates across layers. Comprehensive firing rate metrics and computational costs are detailed in Table~\ref{tab:energy}. Evidently, our SpikeFPN outshines contemporaneous spiking neural networks in automotive object detection tasks, achieving lower firing rates and fewer operations.

Adopting methodologies of energy consumption calculation from~\cite{DBLP:journals/tnn/RathiR23, DBLP:conf/nips/CheLZZMCGL22}, SNN addition operations are quantified using \(s\times T\times A\), where \(s\) signifies the average of the neuron firing rates of all time steps, \(T\) represents the time step, and \(A\) accounts for additions per iteration. Energy estimations are predicated on~\cite{DBLP:conf/isscc/Horowitz14}'s examination of \SI{45}{\nano\meter} CMOS technology, as adopted in works like~\cite{DBLP:journals/tnn/RathiR23, DBLP:conf/nips/LiGZDHG21, DBLP:conf/nips/CheLZZMCGL22}.
Take results from~\cite{DBLP:journals/tnn/RathiR23, DBLP:journals/neuromorphic/KimCP22} as the reference values, SNN addition operations cost \SI{0.9}{\pico\joule}, whereas ANN multiply-accumulate (MAC) operations demand \SI{4.6}{\pico\joule}. We benchmarked against Events-RetinaNet~\cite{DBLP:conf/nips/PerotT0MS20} given its accessible codebase, facilitating a balanced comparison. Notably, compared to Events-RetinaNet, SpikeFPN demonstrates a nearly 6-fold reduction in operations and a 33-fold decrease in energy consumption. Such frugality underscores the SNN's potential in energy-efficient, event-driven vision tasks.

\begin{table}[t]
\caption{Comparison of operation number, firing rate and estimated energy cost. ``\#OP." denotes the number of operations, which is accumulated if the model is an SNN and multiply-accumulate if the model is an ANN.}
\label{tab:energy}
\centering
\setlength{\tabcolsep}{6.5pt}
\renewcommand{\arraystretch}{1.2}
\begin{tabular}{l|c|c|c}
\hline
\textbf{Model} & \textbf{\#OP.} (G) & \textbf{Firing Rate} & \textbf{Energy} (\SI{}{\milli\joule}) \\ 
\hline
Events-RetinaNet (ANN) & 18.73  & - & 86.16  \\
\hline
VGG-11 + SDD & 12.30 & 22.22\% & 11.07  \\
\hline
MobileNet-64 + SSD & 6.39 & 29.44\% & 5.74  \\
\hline
DenseNet121-24 + SSD & 4.33 & 37.20\% & 3.90  \\
\hline
SpikeFPN (ours) & \textbf{2.83} & \textbf{19.10\%} & \textbf{2.55}  \\
\hline
\end{tabular}
\end{table}

\section{Conclusions}
The presented research introduces the SpikeFPN, a novel architecture that seamlessly combines a spiking feature pyramid network with a self-adaptive spiking neuron. This design specifically targets event-based automotive object detection tasks by harnessing the unique capability of the self-adaptive spiking neurons to encode sparse data effectively. The adoption of such architecture, an area of burgeoning interest, has delivered impressive outcomes in terms of both detection accuracy and computational efficiency.

Extensive experimentation using the GEN1 Automotive Detection dataset underscores the efficiency and validity of our proposed model. These experimental results depict not only a commendable mAP prediction accuracy, but also the sound reasoning behind the SpikeFPN's design. In drawing comparisons with similar network structures, inclusive of non-adaptive and various downsampling configurations, our SpikeFPN demonstrates consistent robustness across varying problem sizes. Theoretical insights further reveal the benefits of the spiking mechanism-based model, particularly highlighting its low power consumption and reduced computational demands, suggesting its potential utility for applications demanding efficient computational paradigms.

Notably, although the integration of self-adaptive spiking mechanisms within the feature pyramid network structure presents a promising avenue for advancing event-based visual tasks, there are still some limitations in the current model design.
On the one hand, the current model design relies on the acquisition capability of the data sensors, which may prevent the model from achieving the desired effect when acting on other sensors, resulting in a model that does not satisfy the cross-sensor universality.
On the other hand, in the current data preprocessing and model design, although the potential sparsity can be explored, the interface from data to model is still only a simple frame-pressing method, leaving a gap between the sparse data and the sparse model.
The question of how to fully exploit the generality of models and how to effectively utilize the adaptability of models and data remains to be solved.
Given the inherent discrete and sparse nature of events and the added advantage of preserving temporal information, aligning this data format with our unique neuron setup might pave the way for notable performance enhancements. The inherent low power consumption attributed to the sparse propagation properties of spiking neurons, when combined with our experimental outcomes, suggests a possible edge that SNNs could hold over traditional architectures in specific domains. Such potential merits further exploration and research.

\bibliographystyle{IEEEtran}
\bibliography{main}

\end{document}